\documentclass[a4paper,12pt]{extarticle}
\usepackage[T2A]{fontenc}
\usepackage{graphicx}
\usepackage{lineno,hyperref}
\usepackage[russian, english]{babel}
\usepackage{wrapfig}
\usepackage{amssymb}
\usepackage[a4paper,total={16.5cm, 24.75cm}]{geometry}

\title{Electronic Bursting Neuron: design, equations and hardware implementation} 

\author{Lev V. Takaishvili\textsuperscript{1,2}, Vladimir I. Ponomarenko\textsuperscript{2,3},\\ Maksim V. Kornilov\textsuperscript{1,2}, Ilya V. Sysoev\textsuperscript{1,2}}

\date{
\textsuperscript{1}Peter the Great St. Petersburg Polytechnic University, Russia;\\
\textsuperscript{3}Saratov State University, Russia;\\
\textsuperscript{2}Saratov Branch of Kotelnikov Institute of Radioengineering and Electronics of RAS, Russia
}

\begin{document}

\maketitle

\begin{abstract}
    Electronic neurons are a keystone for construction of the spiking neural networks which have numerous applications in neuroprosthetics, artificial memory, intensive calculations etc. A number of concepts of electronic neurons has been already proposedm with some of them implemented in hardware. However, new schemes are of significant interest since the existing ones do not fit all requirements: either they are too complex and expensive in realization, or they are not able to demonstrate all demanded regimes, or their do not have a appropriate mathematical description and therefore may be investigated only experimentally etc.  
    
    In this study we propose a new design of bursting electronic neuron constructed as a circuit implementation of the equations of a phase-locked loop system. To succeed, we use a novel hybrid approach: we start from the phenomenological equations providing the demanded, then we adjust and modify these equations to simplify the implementation rather than implementing the biophysical equations into thee hardware directly or writing equations for the already constructed circuit. The resulting circuit is simple in implementation and well matches the underlying equations. It can be used for description of not only a single neuron, but small neural circuits too.
\end{abstract}

\section{Introduction}
Modeling dynamic processes in neural systems remains one of the central tasks of modern computational neuroscience and nonlinear dynamics. Classical models such as the Hodgkin-Huxley model \cite{HH1952} and the FitzHugh-Nagumo model \cite{FitzHugh_BiophysJ1961,Nagumo_etal_ProcIREJ1962} laid the foundation for understanding the mechanisms of nerve impulse generation and transmission. At the same time, the researchers are still looking for new physically realizable models capable of reproducing complex modes of neural activity, including chaotic and periodic oscillations \cite{Izhikevich_2007book,Chua_2019HandbookMemristorNet}.
Electronic analogues of neurons are of particular interest. They can be implemented as hardware circuits for creating neuromorphic systems and hardware neural networks \cite{Baryczkowski_2023Electronics,Peng_2024Chip}. This approach allows us to explore the dynamics of neurons in real time and overcome the computational limitations of purely software simulations \cite{Tanaka_NN2019,Markovich_Nature2020}.
Switching from abstract mathematical models to the real world implementations is especially important for tasks of synchronization, signal processing, and studying the collective dynamics of neural ensembles \cite{Larger_OpticsExpress2012,Indiveri_Frontiers2011}. 

A number of hardware implementations of the the electronic concept have been already proposed, with most of them realizing the idea of direct modeling the biophysical equations for transmembrane current starting from the pioneering works by \cite{MahowaldDouglas_Nature1991,RascheDouglas_SignalProc2000,vanSchaik_NeuralNetworks2001} and further implementations of FitzHugh--Nagumo model \cite{Binczak_etal_ElectronLett2003,Li_etal_NonlinearDynamics2012,Kulminskiy_etal_ND2019,Egorov_etal_CSF2022}, Morris--Lecar model \cite{Hu_etal_ND2016} and different, mostly reduced and simplified versions of Hodgkin--Huxley model \cite{FarquharHasler_IEEE_CASI2005,Ma_etal_2012,Rutherford_etal_AmericanJPhys2020}. 

An alternative way is to find a mathamatical model or develop an electronic circuit, which demonstrates spiking and bursting phenomenologically without clear biophysical background. Here, we construct a hardware generator based on the ideas from \cite{MischenkoAND2012,Matrosov_EPhJST2013,Mishchenko_etal_TPhL2017}. In the paper by \cite{MischenkoAND2012} the authors found the neuron-like modes in the equations of a generator proposed much earlier by \cite{Shalfeev1968}. The equations described a model of a phase-locked loop system with three dynamic variables, one of which --- the voltage $y$ might be matched with a trans-membrane voltage. The hardware implementation of this system \cite{Mishchenko_etal_TPhL2017} is more complex, containing, in addition to these equations, a voltage-controlled generator (VCG) and a phase detector.

In this work, we follow the idea from the paper by \cite{MischenkoAND2012}, implementing the circuit with methods of analog modeling of electronic devices. The direct implementation of this concept was not possible since the original model from \cite{Shalfeev1968,MischenkoAND2012} inherits the design of phase-locked loop system, one of its variables --- the phase $\phi$ may grow infinitely. This unlimited growth of the phase variable is a primary problem of a direct hardware implementation of the equations. In the experiment, the phase is usually considered in the interval from 0 to $2\pi$. The second problem is that the equations contain a nonlinear periodic cosine function, which is difficult to construct. To overcome these two problems, a new method for limiting the phase and replacing the nonlinear cosine function with a simpler hyperbolic tangent function is proposed. 
The aim of this work is to develop, mathematically justify, and experimentally test a new analog neuron model.

\section{Mathematical model}
The original model from \cite{MischenkoAND2012} is written in Eq.~(\ref{eq:original})
\begin{eqnarray} \label{eq:original} 
\dot{\phi}(t)&=&y(t), 
 \nonumber\\ 
 \dot{y}(t)&=&z(t), \\ 
 \varepsilon_1 \varepsilon_2 \dot{z}(t)&=&\gamma - (\varepsilon_1+\varepsilon_2)z(t)-(1+\varepsilon_1 cos(\phi))y(t).  \nonumber 
\end{eqnarray}
Equation (\ref{eq:original}) is a reduced equation of a phase-locked loop, and the full circuit it describes is a phase-locked loop that includes a reference frequency generator with a phase detector and feedback. Equation (\ref{eq:original}) does not consider the dynamics of the reference generator, but only works with the phase. 

In this study, we suggest to start from the equations (\ref{eq:original}) rather than from any implementation of phase-locked loop system and to construct an electronic generator following the principles of analog modeling \cite{Ulmann_2023book}. The idea of such a way of modeling is that each first-order equation is associated with an analog integrator. So, in this paradigm, we do not need a phase detector or a voltage-controlled oscillator.

When implementing an electronic circuit according to equation (\ref{eq:original}), the main problem is that the variable $\phi$ --- the phase, increases infinitely. In the computer calculations this is mostly not a problem, but it is impossible to implement in the experiment due to the limited voltage range in the electronic circuit. This issue prevented a direct implementation of the model (\ref{eq:original}) for a long time. One solution is to reset the phase to zero when it reaches $2\pi$. In the experiment, the variable $\phi$ would be reset to ground by an analog switch when it reaches $2\pi$. This would allow the phase to change from 0 to $2\pi$. In calculations, we do not perform this reset because $\cos(\phi)$ is a $2\pi$-periodic function. The variables $y$ and $z$ are bounded and do not require a reset.

\begin{figure}
	\centering
	\includegraphics[width=\linewidth]{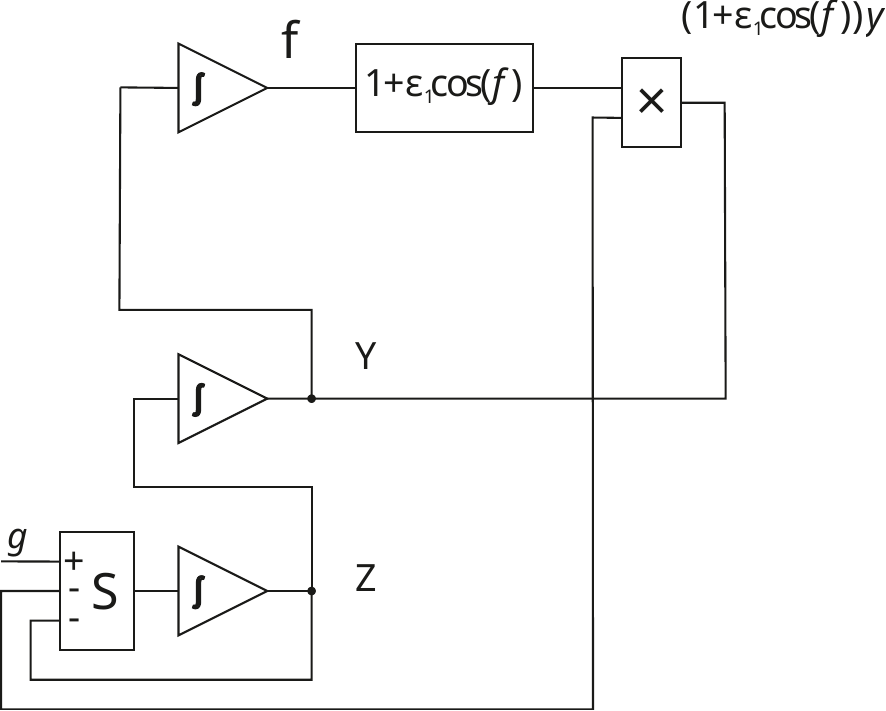}
	\caption{The flowchart corresponding to the equation (\ref{eq:original}).} 
	\label{fig:Block-scheme}
\end{figure}

If we ignore the problem of infinite growth of the variable $\phi$, we can construct a block diagram of an experimental setup using integrators and nonlinear amplifiers from the equations (\ref{eq:original}), as shown in Figure~\ref{fig:Block-scheme}.
This circuit will work as long as $\phi$ is within the operating voltage range of the electronic circuit (close to the supply voltage). Therefore, it is necessary to ensure that $\phi$ is finite in the experiment. To achieve this, the circuit includes a reset mechanism that resets $\phi$ when it reaches a value of $2\pi$. When this threshold is reached, the comparator activates and switches the analog key to a conducting state, causing the output voltage of the integrator to reset to zero.

\begin{figure}
	\centering
	\includegraphics[width=\linewidth]{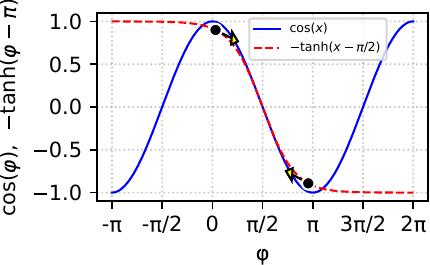}
	\caption{comparison of the functions $\cos(\phi)$ (blue line) and $-\tanh(\phi-\pi/2)$ (red dashed line).}
	\label{fig:cos-tanh}
\end{figure}

In addition, to make this circuit work, one needs to implement a bias voltage that sets $\gamma$, a multiplier, and a nonlinear transformation block $\alpha+\beta \cos(\phi)$, which is the most difficult and costly for implementation in an electronic circuit since it contains a periodic cosine function. Even if we take a section from 0 to $2\pi$, the function $\cos(\phi)$ is non-monotonic, decreasing and then increasing. The change in $\phi$ from 0 to $2\pi$ is shown by arrow 1 in Figure \ref{fig:cos-tanh}. It can be approximated by a polynomial, but of a fairly high order.
Let us note that the cosine function is not only periodic but also symmetric, meaning that if we mirror half a period of $\cos(\phi)$ from $\pi$ to $2\pi$ and superimpose it on the interval from 0 to $\pi$, these two functions completely match. Thus, instead of integrating from 0 to $2\pi$ one can integrate from 0 to $\pi$, and then change the sign and integrate it back to 0. Moreover, it is possible to change the sign of the variable $y$ instead of changing the sign of $\phi$, and then $\phi$ will start decreasing without a gap, with the value of the function $\cos(\phi)$ exactly matching.

Thus, the algorithm for the circuit operation may be verbally described as follows. One integrates the variable $y$ to obtain the variable $\phi$ from 0 to $\pi$. When the phase reaches $\pi$, $y$ is switched to $-y$ in the equation for $\dot{\phi}$, continuing the integration until it reaches 0. When $y$ reaches 0, $-y$ is switched back to $y$, and so on. Note that in this case it is necessary to implement a fairly simple function approximating the cosine in the range from 0 to $\pi$. We suggest using the function $-\tanh(\phi-\pi/2))$, which is better suited for implementation in the experiment. Both functions $\cos(\phi)$ and $\tanh(\phi-\pi/2))$ are plotted in Fig. \ref{fig:cos-tanh}. They match well for a range between yellow arrows which show the direction of $\phi$ change, corresponding to $y$ or $-y$ term in the modified equations (\ref{eq:oscillator-pi2}).
\begin{eqnarray} \label{eq:oscillator-pi2} 
\dot{\phi}(t)&=&y(t), \textrm{ if } \phi \to \leqslant -\frac{\pi}{2};  
-y(t), \textrm{ if } \phi \to \geqslant \frac{\pi}{2} \nonumber\\
\dot{y}(t)&=&z(t), \\ 
 \varepsilon_1 \varepsilon_2 \dot{z}(t)&=&\gamma - (\varepsilon_1+\varepsilon_2)z(t) - \left(1 + \varepsilon_1 \tanh\left(\phi-\frac{\pi}{2}\right)\right) y(t).  \nonumber 
\end{eqnarray}

\begin{figure}
	\centering
	\includegraphics[width=\linewidth]{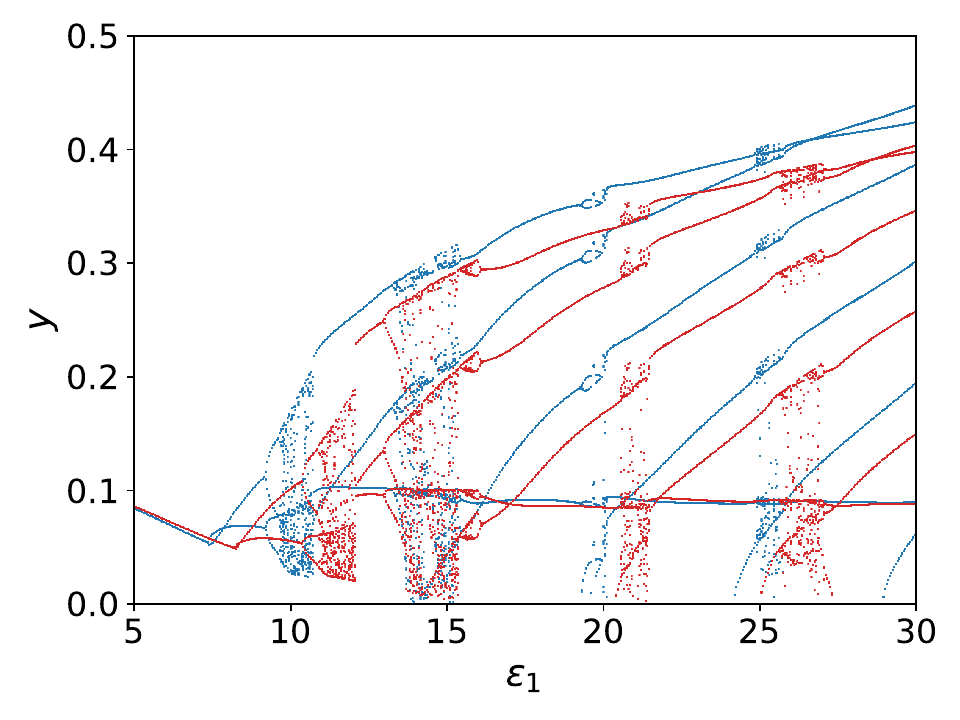}
	\caption{Bifurcation diagrams for the parameter $\varepsilon_1$ at $\varepsilon_2=10$, $\gamma=0.25$. The diagram for the original equations (\ref{eq:original}) is plotted in blue and the diagram for the modified equations (\ref{eq:oscillator-pi2}) is plotted on red. The bifurcation diagrams were constructed using the section of the phase space by the plane $\phi = \pi/4$.}
	\label{fig:bif_diargam}
\end{figure}

To compare the dynamical regimes of the original model (\ref{eq:original}) and the modified one (\ref{eq:oscillator-pi2}), we constructed one-dimensional bifurcation diagrams of  with fixed parameters $\varepsilon_2=10$, $\gamma=0.25$, and varying $\varepsilon_1$ from $5$ to $30$. Since the model (\ref{eq:oscillator-pi2}) has a discontinuous function at the right side of the first equation (and since its evolution operator actually depends on events) the integration was performed using the 2nd-order Runge-Kutta method with a step size of $10^{-4}$, see Fig.~\ref{fig:bif_diargam}. Calculations were performed without inheritance using the same initial conditions for all parameter values and skipping the transient of 4000 time units ($4 \cdot 10^7$ data points). Series of length 1000 dimensionless time units were used to construct a map (with additional interpolation used to improve the precision) by using a section of the phase space with a plane $\phi = \pi/4$. In the original paper by \cite{MischenkoAND2012} the authors used a plane $\phi = \pi$, but it is not applicable in our case since such values of $\phi$ are unreachable for the model (\ref{eq:oscillator-pi2}).

The Fig.~\ref{fig:bif_diargam} shows that the bifurcation diagrams match each other qualitatively. All regimes present in the original system (\ref{eq:original}) also remain in the modified one (\ref{eq:oscillator-pi2}), but with somewhat different values of the parameter $\varepsilon_1$.

\section{Analog generator}
In an experimental hardware device all variables and parameters must actually be dimensional. So, let us modify the equations (\ref{eq:oscillator-pi2}) by renormalization. First, let us introduce the dimensional time using the $RC$ time constant (it has a dimension of time and determines the circuit frequency). Also, let us introduce new parameters, more convenient for practical use and implementation: $\varepsilon={\varepsilon_1 \varepsilon_2}$, $\Gamma=\gamma$, $\alpha=(\varepsilon_1+\varepsilon_2)$ and $\beta={\varepsilon_1}$. The variables $\phi$, $y$, $z$ are measured in volts and the equation (\ref{eq:oscillator-pi2}) is rewritten as (\ref{eq:oscillator-new}).
\begin{eqnarray} \label{eq:oscillator-new} 
RC \dot{\phi}(t)&=&y(t), \nonumber\\ 
RC \dot{y}(t)&=&z(t), \\ 
\varepsilon RC\dot{z}(t)&=&\gamma - {\alpha}z(t) - y(t) - \beta \tanh(\phi-\pi/2)) y(t).  \nonumber 
\end{eqnarray}

\begin{figure}
	\centering
	\includegraphics[width=\linewidth]{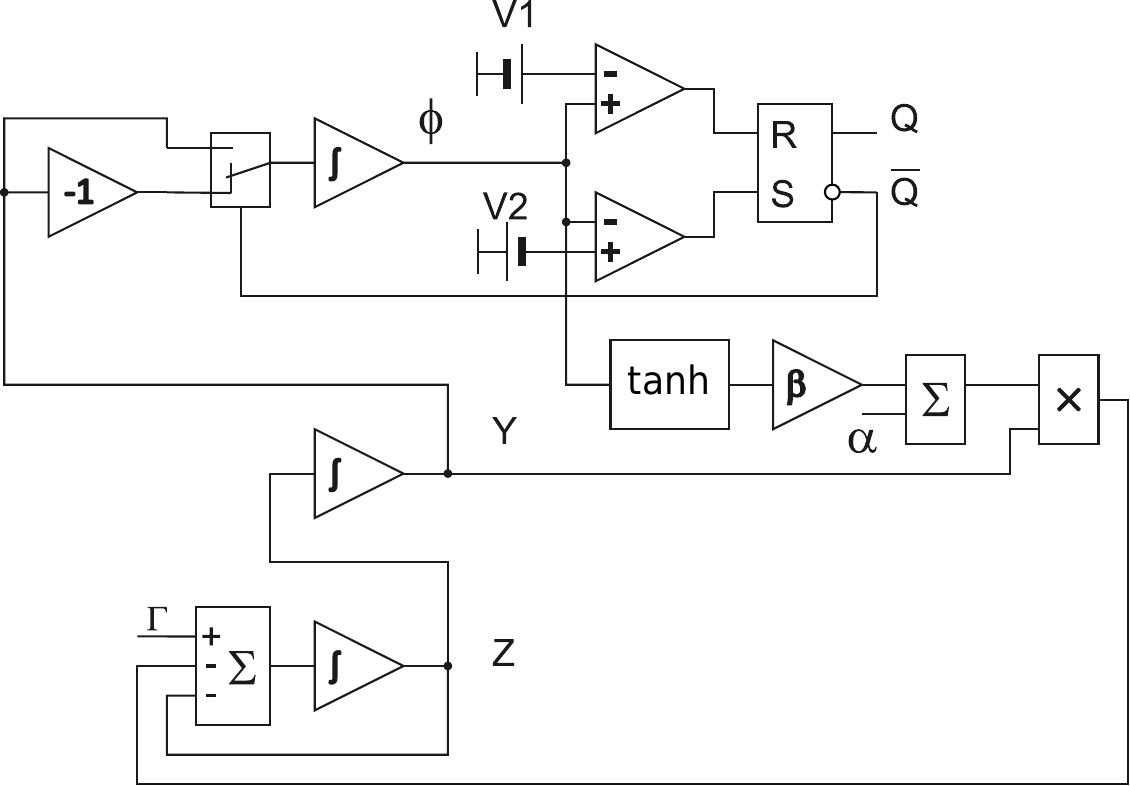}
    \caption{Block diagram of the modified electronic neuron model described by the equations (\ref{eq:oscillator-new}).}
	\label{fig:diagram-new}
\end{figure}
To control the variable $\phi$, two comparators, an RS-trigger, an analog switch S$_1$, and an amplifier with a gain of $-1$ are included in the circuit. Let the RS-trigger be set when the power is turned on; then the S$_1$ switch is in the upper position, and the variable $\phi$ increases. When the value of $\phi$ reaches $\frac{\pi}{2}$, the upper comparator activates and resets the trigger. The S$_1$ switch is turned on, and the value $-y$ is fed to the input of the upper integrator instead of $y$, causing the variable $\phi$ to decrease. When the value of $\phi$, in turn, reaches $-\frac{\pi}{2}$, the switch S$_1$ is turned off, and the value $y$ is fed to the input of the upper integrator, causing the variable $\phi$ to increase once again.

\begin{figure}
	\centering
	\includegraphics[width=\linewidth,angle=180]{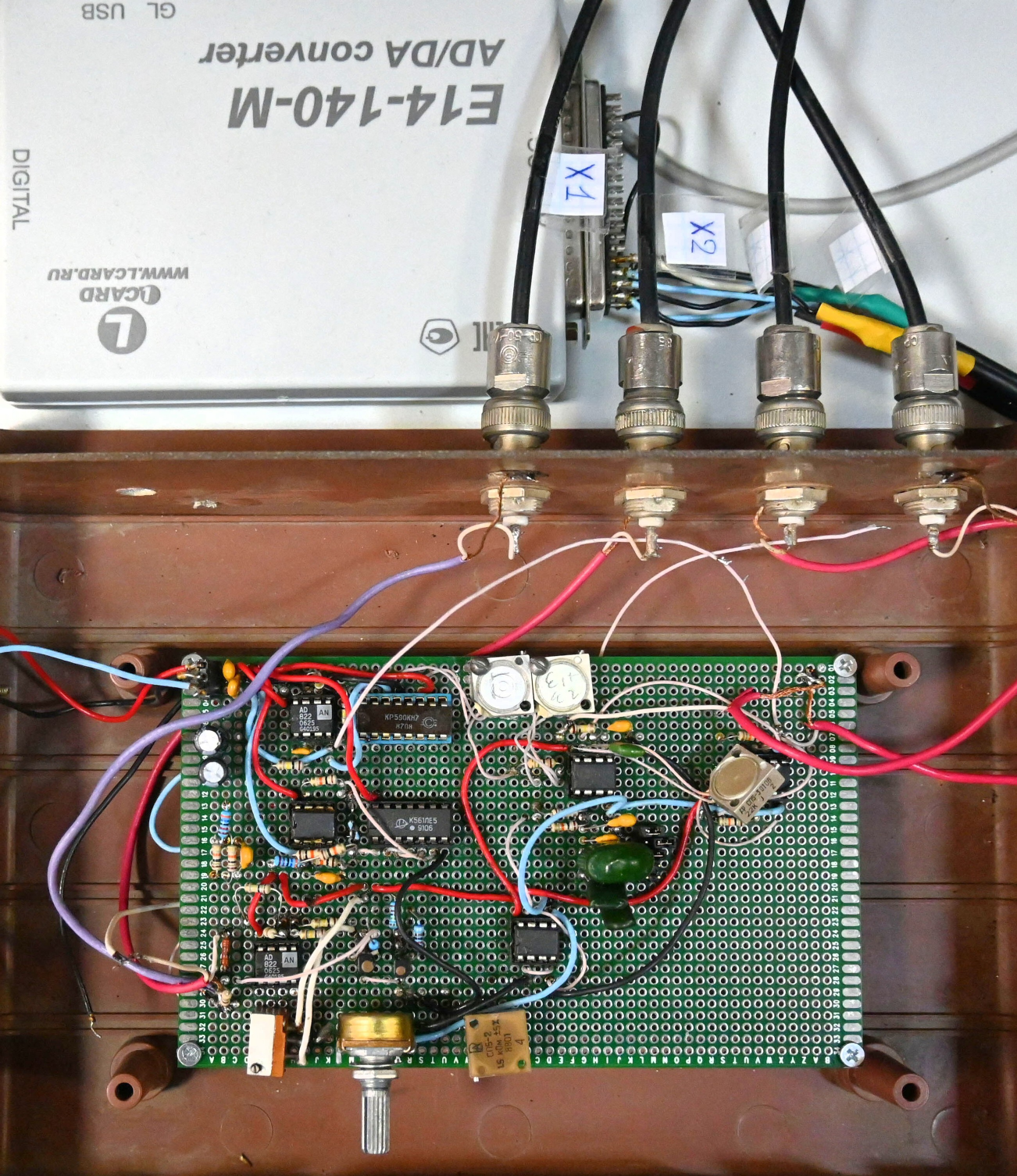}
	\caption{The photo of the developed and constructed hardware device.}
	\label{fig:photo}
\end{figure}

\begin{figure}
	\centering
	\includegraphics[width=\linewidth]{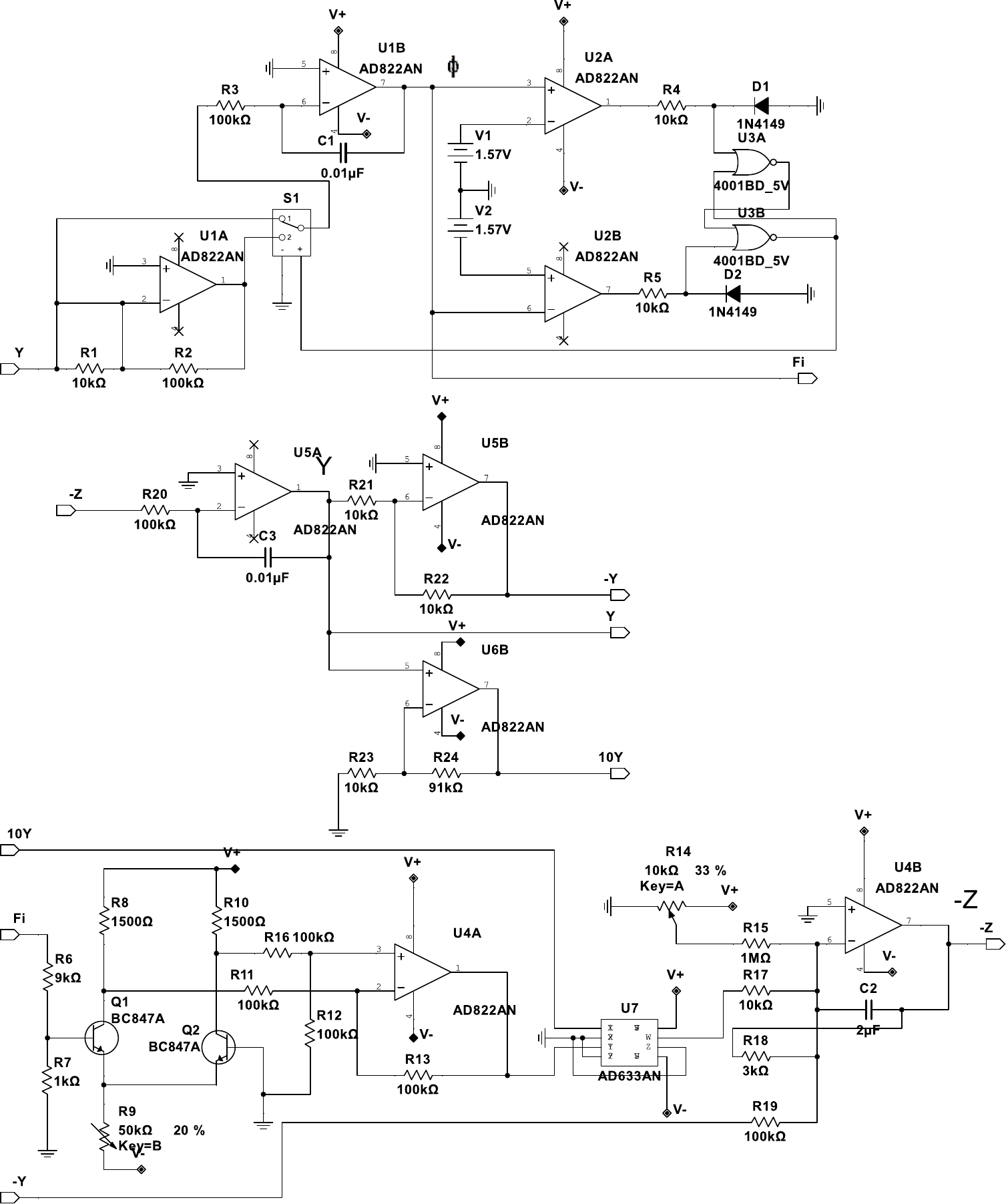}
    \caption{The engineering constructive scheme of the modified electronic neuron model described by the equations (\ref{eq:oscillator-new}) implementing the block diagram from Fig.~\ref{fig:diagram-new} using actual electronic elements.}
	\label{fig:engineering}
\end{figure}

Based on the block diagram in Fig.~\ref{fig:diagram-new}, a hardware circuit was constructed, see the photo in Fig.~\ref{fig:photo}. Note that the operational amplifier based integrators invert an output since the output voltage is proportional to the negative integral of the input voltage. The engineering constructive scheme implementing the block diagram \ref{fig:diagram-new} is plotted in fig.~\ref{fig:engineering} with actual electronic components used for construction.

The integrator for the variable $\phi$ is made using the operational amplifier $U1B$, the capacitor $C1$ and the resistor $R1$. The input is either variable $y$ or $-y$, with switching provided by the key $S1$. The variable $\phi$ is supplied to the inputs of two comparators: $U2A$, $U2B$, the thresholds of which are set equal to $+1.57 V$ and $-1.57 V$ respectively. When the threshold of $+1.57 V$ is reached, the comparator $U2A$ is triggered and the $RS$-triggers $U3A$, $U3B$ are reset, the key $S1$ is switched to the lower position, and the variable $\phi$ begins to decrease. When the threshold of $-1.57 V$ is reached, the comparator $U2B$ is triggered, and the key $S1$ is switched to the upper position. The variable $\phi$ begins to increase.

The variable $\phi$ is fed to the input of  $z$ integration circuit. Here, the transistors $Q1$, $Q2$, and the operational amplifier $U4A$ form the hyperbolic tangent function, and the analog multiplier forms the product $\beta \tanh(\phi) y(t)$. The parameter $\beta$ is adjusted using the potentiometer $R9$. The parameter $\Gamma$ is controlled by the resistor $R14$. The integrator $U4B$ forms the variable $z$. The variable $y$ is the integral of $z$ according to the equation, and the integrator $U5A$ is used to form it.

\begin{figure}[tp]
	\centering
	\includegraphics[width=0.75\linewidth]{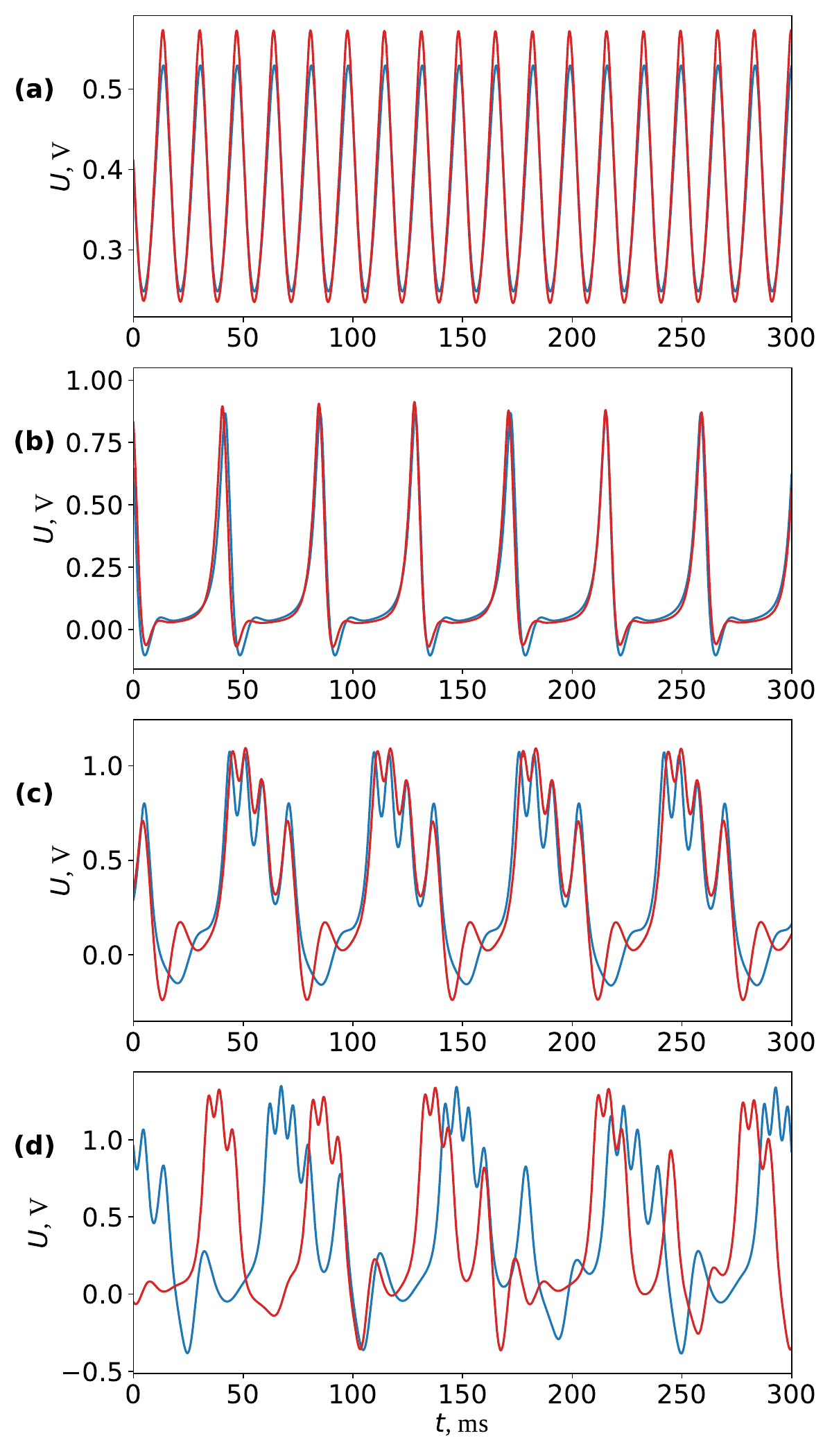}
    \caption{Time series of the mathematical model (blue curves) and the experimental device (red curves) in different regimes: (a) --- quasilinear regime ($\gamma=0.3727$, $\varepsilon=20$, $\alpha=12$, $\beta=2$), (b) --- spiking regime ($\gamma=0.145$, $\varepsilon=18$, $\alpha=10$, $\beta=6$), (c) --- periodic bursting regime ($\gamma=0.38$, $\varepsilon=78$, $\alpha=15$, $\beta=14.5$), (d) --- chaotic bursting regime ($\gamma=0.39$, $\varepsilon=78$, $\alpha=10$, $\beta=14.8$).}
	\label{fig:series}
\end{figure}

In accordance with the above scheme, see Fig.~\ref{fig:engineering}, an experimental setup was manufactured, and the dynamics of the mathematical model and the experimental setup were compared. Both experimental and model time series are plotted in Fig.~\ref{fig:series}.
One can see that the experimental and numerical series match each other well in all regimes, though some shape and amplitude difference is present. Please note that there is a fundamental inaccuracy in setting the parameters of the experimental generator. This inaccuracy is a result of generally acceptable inaccuracies in the nominal values of the elements as well as inaccuracy of setting resistances using the potentiometer $R9$.
 
\section{Conclusion and Discussion}
In this paper, a new analog neuron model constructed as a schematic realization of modified phase-locked loop equations \cite{Shalfeev1968,MischenkoAND2012} was proposed and studied. For the convenience of hardware implementation, the nonlinear periodic cosine function from the original model (\ref{eq:original}) was replaced by a hyperbolic tangent function in the modified model (\ref{eq:oscillator-new}), with the new equations containing a discontinuous function in the first equation, experimentally implemented using a comparator, RS-triggers and a switch. This allowed us to avoid implementing the periodic $\cos$ function in the experiment and solve the problem of unlimited  growth of the phase variable $\phi$. We constructed a hardware implementation of the proposed system and studied it varying parameters.

Then, we compared the dynamical regimes of the proposed model with those of the original one, following the sections of the parameter space published by \cite{Matrosov_EPhJST2013}. Since the modified system has limited range of $\phi$, we switched to somewhat different sections parallel to those used by \cite{Matrosov_EPhJST2013} and performed calculations both for the original system and for the modified one. In addition we made a series of experimental observations for some particular parameter values. As a result, we showed that all types of dynamical modes found in the original system, including different periodic and chaotic bursting, may be detected both in the modified mathematical model and in the experiment. However, due to the difference in nonlinear functions same modes are realized for somewhat different values of parameters, with this difference being systematical: similar regimes in the modified system appear for larger values of inertia parameter. 

Comparing the suggested generator to the already known electronic models of bursting neurons, we have to say that its first advantage over the original system \cite{MischenkoAND2012} is that the experimental setup matches much better the equations than for the original system, see \cite{MischenkoAND2012,Mishchenko_etal_TPhL2017}. However, to characterized to which extent the mathematical model describes the experimental setup, wee need to perform the reconstruction from time series like it was done for the original system by \cite{Mishchenko_etal_CASII2022}.
Good correspondence between the original and modified systems means that most results of computational studies of equations should be easily translated to the experimental setup. This issue is of the significant importance for any complex modes, especially found in the neuron ensembles, including dynamics out of attractor \cite{vanOoytien_1992DiolCybernetics,Riecke_Chaos2007,Afraimovich_IJBCh2004} which may be useful for description of complex oscillatory regimes of finite duration like epileptic seizures \cite{Egorov_NonlinearDyn2024}. At the same time, the proposed scheme contains minimal number of equations necessary for bursting and chaotic behavior; for a comparison see e.\,g. \cite{SavinoFormigli_Biosystems2009} where another bursting electronic neuron model described by four equations was constructed and studied. 

\subsection*{Acknowledgments}
The research was supported by Russian Science Foundation, Russia (project No. 25-12-00176, \texttt{https://rscf.ru/project/25-12-00176/}).
	
\bibliography{NeuronVIP}

@article{FitzHugh_BiophysJ1961,
  author={FitzHugh, R},
  title={Impulses and physiological states in theoretical models of nerve membrane},
  journal={Biophysical J.},
  volume={1},
  pages={445--466},
  year={1961}
}

@article{Nagumo_etal_ProcIREJ1962,
  author={Nagumo, J. and Arimoto, S. and Yoshizawa, S.},
  title={An active pulse transmission line simulating nerve axon},
  journal={Proc. IRE},
  volume={50},
  pages={2061--2070},
  year={1962}
}

@article{MahowaldDouglas_Nature1991,
	author  = {Misha Mahowald and Rodney Douglas},
	title   = {A silicon neuron},
	journal = {Nature},
	year    = {1991},
	volume  = {354},
	pages   = {515--518},
	doi = {10.1038/354515a0}
}

@article{RascheDouglas_SignalProc2000,
	author  = {Rasche, C. and Douglas, R.},
	title   = {An Improved Silicon Neuron},
	journal = {Analog Integrated Circuits and Signal Processing},
	volume  = {23},
	pages   = {227–236},
	year    = {2000},
	doi     = {10.1023/A:1008357931826},
}

@article{Binczak_etal_ElectronLett2003,
	title   = {Experimental study of bifurcations in modified {FitzHugh--Nagumo} cell},
	author  = {Binczak, S and Kazantsev, V B and Nekorkin, V I and Bilbault, J M},
	journal = {Electronics Letters},
	volume  = {39},
	number  = {13},
	pages   = {1},
	year    = {2003},
	doi     = {doi:10.1049/el:20030657},
}

@article{Shalfeev1968,
	author   = {Shalfeev, V. D.},
	title    = {Investigation of the dynamics of a system of automatic phase control of frequency with a coupling capacitor in the control loop},
	journal  = {Radiophys Quantum Electron},
	year     = {1968},
	volume   = {11},
	number   = {3},
	pages    = {221–226},
}

@article{Mishchenko_etal_TPhL2017,
	author   = {Mishchenko, M. A. and Bolshakov, D. I. and Matrosov, V.V.},
	title    = {Instrumental implementation of a neuronlike generator with spiking and bursting dynamics based on a phase-locked loop},
	journal  = {Tech. Phys. Lett.},
	year     = {2017},
	volume   = {43},
	pages    = {596–599},
	doi      = {10.1134/S1063785017070100},
}

@ARTICLE{Mishchenko_etal_CASII2022,
	author  = {Mishchenko, Mikhail A. and Bolshakov, Denis I. and Vasin, Alexander S. and Matrosov, Valery V. and Sysoev, Ilya V.},
    journal = {IEEE Transactions on Circuits and Systems II: Express Briefs},
    title   = {Identification of Phase-Locked Loop System From Its Experimental Time Series},
    year    = {2022},
    volume  = {69},
    number  = {3},
    pages   = {854-858},
    doi     = {10.1109/TCSII.2021.3122892},
}

@article{MischenkoAND2012,
	author   = {Mishchenko, M. A. and Shalfeev, V. D. and Matrosov, V. V.},
	title    = {Neuron-like dynamics in phase-locked loop},
	journal  = {Izvestiya VUZ. Applied Nonlinear Dynamics},
	year     = {2012},
	volume   = {20},
	number   = {4},
	pages    = {122--130},
	doi      = {10.18500/0869-6632-2012-20-4-122-130},
}

@article{HH1952,
  author		= "Hodgkin, A. and Huxley, A.",
  title			= "A quantitative description of membrane current and its application to conduction and excitation in nerve",
  journal		= "J. Physiol.",
  volume		= "117",
  pages			= "500--544",
  year			= "1952",
  doi			= "10.1113/jphysiol.1952.sp004764"
}

@book{Izhikevich_2007book,
  title   = {Dynamical systems in neuroscience},
  author  = {Izhikevich, Eugene M},
  year    = {2007},
  publisher = {MIT press},
  address = {Cambridge, MA, USA},
  doi     = {},
}

@article{Chua_2019HandbookMemristorNet,
  title   = {Memristor, Hodgkin-Huxley, and edge of chaos},
  author  = {Chua, Leon},
  journal = {Handbook of Memristor Networks},
  pages   = {287--313},
  year    = {2019},
  doi     = {10.1007/978-3-319-76375-0\_10},
}

@article{Peng_2024Chip,
  title   = {Memristor-based spiking neural networks: cooperative development of neural network architecture/algorithms and memristors},
  author  = {Peng, Huihui and Gan, Lin and Guo, Xin},
  journal = {Chip},
  volume  = {3},
  number  = {2},
  pages   = {100093},
  year    = {2024},
  doi     = {10.1016/j.chip.2024.100093},
}

@article{Baryczkowski_2023Electronics,
  title   = {Study of the complexity of CMOS neural network implementations featuring heart rate detection},
  author  = {Baryczkowski, Piotr and Szczepaniak, Sebastian and Matykiewicz, Natalia and Perz, Kacper and Szcz{\k{e}}sny, Szymon},
  journal = {Electronics},
  volume  = {12},
  number  = {20},
  pages   = {4291},
  year    = {2023},
  doi     = {10.3390/electronics12204291},
}

@article{Tanaka_NN2019,
    author    = {G. Tanaka and T. Yamane and J. B. Héroux and R. Nakane and N. Kanazawa and S. Takeda and H. Numata and D. Nakano and A. Hirose},
    title     = {Recent advances in physical reservoir computing: A review},
    journal   = {Neural Networks},
    volume    = {115},
    pages     = {100-123},
    year      = {2019},
    doi       = {10.1016/j.neunet.2019.03.005},
}

@article{Markovich_Nature2020,
    author    = {D. Marković and A. Mizrahi and D. Querlioz and J. Grollier},
    title     = {Physics for neuromorphic computing},
    journal   = {Nature Reviews Physics},
    year      = {2020},
    pages     = {499-510},
    volume    = {2},
    doi       = {10.1038/s42254-020-0208-2},
}

@article{Larger_OpticsExpress2012,
  title   = {Photonic information processing beyond {Turing}: an optoelectronic implementation of reservoir computing},
  author  = {Larger, Laurent and Soriano, Miguel C and Brunner, Daniel and Appeltant, Lennert and Guti{\'e}rrez, Jose M and Pesquera, Luis and Mirasso, Claudio R and Fischer, Ingo},
  journal = {Optics express},
  volume  = {20},
  number  = {3},
  pages   = {3241--3249},
  year    = {2012},
  doi     = {10.1364/OE.20.003241},
}

@article{Indiveri_Frontiers2011,
  title    = {Neuromorphic silicon neuron circuits},
  author   = {Indiveri, Giacomo and Linares-Barranco, Bernab{\'e} and Hamilton, Tara Julia and Schaik, Andr{\'e} van and Etienne-Cummings, Ralph and Delbruck, Tobi and Liu, Shih-Chii and Dudek, Piotr and H{\"a}fliger, Philipp and Renaud, Sylvie and others},
  journal  = {Frontiers in neuroscience},
  volume   = {5},
  pages    = {73},
  year     = {2011},
  doi      = {10.3389/fnins.2011.00073},
}

@book{Ulmann_2023book,
  title    = {Analog and hybrid computer programming},
  author   = {Ulmann, Bernd},
  year     = {2023},
  publisher={Walter de Gruyter GmbH \& Co KG},
  pages    = {314},
  edition  = {2nd},
}

@article{Matrosov_EPhJST2013,
  title   = {Neuron-like dynamics of a phase-locked loop},
  author  = {Matrosov, Valery V and Mishchenko, Mikhail A and Shalfeev, Vladimir D},
  journal = {The European Physical Journal Special Topics},
  volume  = {222},
  number  = {10},
  pages   = {2399--2405},
  year    = {2013},
  doi     = {10.1140/epjst/e2013-02024-9}
}

@article{vanOoytien_1992DiolCybernetics,
  title     = {The emergence of long-lasting transients of activity in simple neural networks},
  author    = {van~Ooytien, A and Van~Pelt, J and Corner, MA and Lopes~da~Silva, FH},
  journal   = {Biological cybernetics},
  volume    = {67},
  number    = {3},
  pages     = {269--277},
  year      = {1992},
  doi       = {10.1007/BF00204400},
}

@article{Afraimovich_IJBCh2004,
  title   = {Heteroclinic contours in neural ensembles and the winnerless competition principle},
  author  = {Afraimovich, Valentin S and Rabinovich, Mikhail I and Varona, Pablo},
  journal = {International Journal of Bifurcation and Chaos},
  volume  = {14},
  number  = {04},
  pages   = {1195--1208},
  year    = {2004},
  doi     = {10.1142/S0218127404009806},
}

@article{Riecke_Chaos2007,
  title   = {Multiple attractors, long chaotic transients, and failure in small-world networks of excitable neurons},
  author  = {Riecke, Hermann and Roxin, Alex and Madruga, Santiago and Solla, Sara A},
  journal = {Chaos: An Interdisciplinary Journal of Nonlinear Science},
  volume  = {17},
  number  = {2},
  year    = {2007},
  doi     = {10.1063/1.2743611},
}

@article{Egorov_NonlinearDyn2024,
  title   = {Hardware implementation of the ring generator with tunable frequency based on electronic neurons},
  author  = {Egorov, Nikita M and Sysoeva, Marina V and Kornilov, Maksim V and Ponomarenko, Vladimir I and Sysoev, Ilya V},
  journal = {Nonlinear Dynamics},
  volume  = {112},
  number  = {13},
  pages   = {11471--11481},
  year    = {2024},
  doi     = {10.1007/s11071-024-09671-z},
}

@article{SavinoFormigli_Biosystems2009,
    title   = {Nonlinear electronic circuit with neuron like bursting and spiking dynamics},
    author  = {Guillermo V. Savino and Carlos M. Formigli},    
    journal = {Biosystems},
    volume  = {97},
    number  = {1},
    pages   = {9-14},
    year    = {2009},
    doi     = {10.1016/j.biosystems.2009.03.005},
}

@article{vanSchaik_NeuralNetworks2001,
	title   = {Building blocks for electronic spiking neural networks},
	journal = {Neural Networks},
	volume  = {14},
	number  = {6},
	pages   = {617--628},
	year    = {2001},
	doi     = {10.1016/S0893-6080(01)00067-3},
	author  = {A. {van Schaik}},
}

@ARTICLE{Li_etal_NonlinearDynamics2012,
	author  = {Li, F. and Liu, Q. and Guo, H. and Zhao, Y. and Tang, J. and Ma, J.},
	title   = {Simulating the electric activity of {FitzHugh-Nagumo} neuron by using Josephson junction model},
	journal = {Nonlinear Dynamics},
	year    = {2012},
	volume  = {69},
	number  = {4},
	pages   = {2169--2179},
	doi     = {10.1007/s11071-012-0417-z},
}

@article{Kulminskiy_etal_ND2019,
	author  = {Kulminskiy, D.D. and Ponomarenko, V.I. and Prokhorov, M.D. and Hramov, A.E.},
	title   = {Synchronization in ensembles of delay-coupled nonidentical neuronlike oscillators},
	journal = {Nonlinear Dynamics},
	year    = {2019},
	volume  = {98},
	number  = {1},
	pages   = {735--748},
	doi     = {10.1007/s11071-019-05224-x},
}

@article{Egorov_etal_CSF2022,
    title   = {Complex regimes in electronic neuron-like oscillators with sigmoid coupling},
    author  = {Egorov, Nikita M and Sysoev, Ilya V and Ponomarenko, Vladimir I and Sysoeva, Marina V},
    journal = {Chaos, Solitons \& Fractals},
    volume  = {160},
    pages   = {112171},
    year    = {2022},
    doi     = {10.1016/j.chaos.2022.112171},
}

@article{Hu_etal_ND2016,
    title   = {An electronic implementation for {Morris--Lecar} neuron model},
    author  = {Hu, Xiaoyu and Liu, Chongxin and Liu, Ling and Ni, Junkang and Li, Shilei},
    journal = {Nonlinear Dynamics},
    volume  = {84},
    pages   = {2317--2332},
    year    = {2016},
    doi     = {10.1007/s11071-016-2647-y},
}

@article{Rutherford_etal_AmericanJPhys2020,
    title   = {Analog implementation of a Hodgkin--Huxley model neuron},
    author  = {Rutherford, George H and Mobille, Zach D and Brandt-Trainer, Jordan and Follmann, Rosangela and Rosa, Epaminondas},
    journal = {American Journal of Physics},
    volume  = {88},
    number  = {11},
    pages   = {918--923},
    year    = {2020},
    doi     = {10.1119/10.0001072},
}

@article{Ma_etal_2012,
    title   = {Bursting Hodgkin--Huxley model-based ultra-low-power neuromimetic silicon neuron},
    author  = {Ma, Qingyun and Haider, Mohammad Rafiqul and Shrestha, Vinaya Lal and Massoud, Yehia},
    journal = {Analog Integrated Circuits and Signal Processing},
    volume  = {73},
    pages   = {329--337},
    year    = {2012},
    doi     = {10.1007/s10470-012-9888-6},
}

@article{FarquharHasler_IEEE_CASI2005,
  title   = {A bio-physically inspired silicon neuron},
  author  = {Farquhar, Ethan and Hasler, Paul},
  journal = {IEEE Transactions on Circuits and Systems I: Regular Papers},
  volume  = {52},
  number  = {3},
  pages   = {477--488},
  year    = {2005},
  doi     = {10.1109/TCSI.2004.842871},
}

\end{document}